\documentclass[final,5p,times,twocolumn]{elsarticle}

\usepackage{amssymb}
\usepackage{amsmath}
\usepackage{graphicx}
\biboptions{sort&compress}
\usepackage{bm}
\usepackage{booktabs}
\usepackage{stfloats}
\usepackage{array}
\usepackage{graphicx}
\usepackage{subcaption}
\usepackage{float}
\usepackage{booktabs}
\usepackage{threeparttable}

\graphicspath{{image/}}

\usepackage{multirow}
\usepackage{amssymb}
\usepackage{color} 
\usepackage[colorlinks,bookmarksopen,bookmarksnumbered,citecolor=blue, linkcolor=blue, urlcolor=blue]{hyperref}
\usepackage{makecell}
\usepackage{verbatim}
\usepackage{tabularx}

\journal{Optics Laser Technology}

\makeatother

\begin{document}

\begin{frontmatter}

\title{Perceptual Region-Driven Infrared-Visible Co-Fusion for Extreme Scene Enhancement} 

\author[label1,label2]{Jing Tao}

\author[label3,label4]{Yonghong Zong}

\author[label1,label2]{Banglei Guan\corref{cor}}
\ead{banglei.guan@hotmail.com}

\author[label1,label2]{Pengju Sun}

\author[label1,label2]{Taihang Lei}

\author[label1,label2]{Yang Shang}

\author[label1,label2]{Qifeng Yu}

\cortext[cor]{Corresponding author}
\affiliation[label1]{organization={College of Aerospace Science and Engineering, National University of Defense Technology},
            city={Changsha},
            postcode={410073}, 
            state={Hunan},
            country={China}}

\affiliation[label2]{organization={Hunan Provincial Key Laboratory of Image Measurement and Vision Navigation},
            city={Changsha},
            postcode={410073}, 
            state={Hunan},
            country={China}}

\affiliation[label3]{organization={Beijing Institute of Tracking and Telecommunication Technology},
          	city={Beijing},
          postcode={100094}, 
          country={China}}
            
\affiliation[label4]{organization={National Key Laboratory of Space Integrated Information System},
	city={Beijing},
	postcode={100094}, 
	country={China}}
\begin{abstract}
In photogrammetry, accurately fusing infrared (IR) and visible (VIS) spectra while preserving the geometric fidelity of visible features and incorporating thermal radiation is a significant challenge, particularly under extreme conditions. Existing methods often compromise visible imagery quality, impacting measurement accuracy. To solve this, we propose a region perception-based fusion framework that combines multi-exposure and multi-modal imaging using a spatially varying exposure (SVE) camera. This framework co-fuses multi-modal and multi-exposure data, overcoming single-exposure method limitations in extreme environments.
The framework begins with region perception-based feature fusion to ensure precise multi-modal registration, followed by adaptive fusion with contrast enhancement. A structural similarity compensation mechanism, guided by regional saliency maps, optimizes IR-VIS spectral integration. Moreover, the framework adapts to single-exposure scenarios for robust fusion across different conditions. Experiments conducted on both synthetic and real-world data demonstrate superior image clarity and improved performance compared to state-of-the-art methods, as evidenced by both quantitative and visual evaluations.
\end{abstract}


\end{frontmatter}

\section{Introduction}
\label{sec1}

Extreme launch site conditions, marked by haze interference and dynamic light intensity variations, severely degrade image quality and complicate subsequent measurement tasks \cite{2014High,Gao23}. These challenges underscore the need for high-dynamic-range and multi-spectral image fusion technologies specifically designed for such environments. Addressing this need is crucial for enhancing electro-optical imaging systems, improving image clarity and measurement accuracy, and supporting decision-making processes at launch sites.

In recent years, multi-modal image fusion, particularly the integration of infrared (IR) and visible light (VIS) images, has garnered considerable research attention \cite{IBFusion,LDFusion,survey,pixel-level}. IR images capture thermal radiation, which accentuates target features, whereas VIS images provide detailed texture and structural information. The fusion of these modalities broadens the spectral range and improves image clarity. Fusion techniques are generally categorized into traditional methods and deep learning-based approaches. Traditional algorithms, such as IVFusion \cite{IVFusion} and PFF \cite{PFF}, leverage image features to guide the fusion process. IVFusion employs multi-scale transformations and norm optimization to achieve effective fusion in complex scenes, while PFF mitigates cross-modality artifacts through fusion in the visual response space \cite{ZHANG2017,ZHOU201615}.

Deep learning-based methods, in contrast, offer more advanced information representation \cite{IBFusion}. Recently, there has been increasing interest in integrating various processing mechanisms into fusion frameworks \cite{TANG202228, MMIF, MURF}. For example, CDDFuse \cite{MMIF} uses feature decoupling, combining CNNs and Transformers to produce fused images that preserve the strengths of both modalities. Wang \emph{et al.} \cite{LDFusion} incorporated natural language processing to improve fusion outcomes, circumventing the limitations of traditional modeling. However, in specialized scenarios such as launch sites, the limited availability of training data hinders the effective training of deep learning models, reducing the generalizability of these methods.

\begin{figure}[tp]
	\centering
	\includegraphics[width=0.47\textwidth]{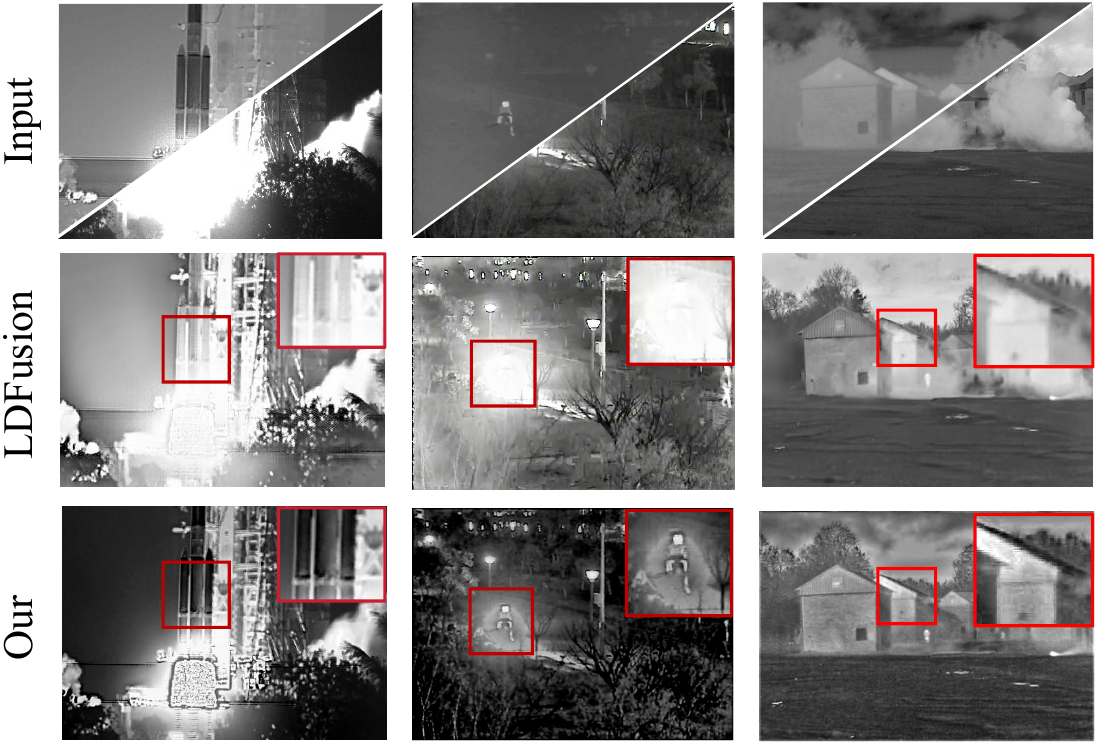}
	\caption{Schematic illustration of different multi-modal image fusion tasks (first row: source images; second row: fusion results using a state-of-the-art method (LDFusion \cite{LDFusion}); third row: our fusion results).}
	\label{fig:taste}
\end{figure}

Currently, existing methods primarily involve the fusion of single-exposure visible light images and infrared images, most of which utilize full pixel integration. Although more abundant information is obtained, the low resolution and noise of the infrared itself are not considered. Therefore, although the obtained results have a good visualization effect, they lose their measurement-related features, which is very unfavorable for subsequent measurement tasks. In order to solve this limitation, we try a complementary fusion idea to integrate additional infrared information on the premise of enhancing and retaining the original information of visible light images. This can achieve the maintenance of target measurement features and structural integrity in extreme scenarios.

\begin{figure*}[tp]
	\centering
	\includegraphics[width=0.95\textwidth]{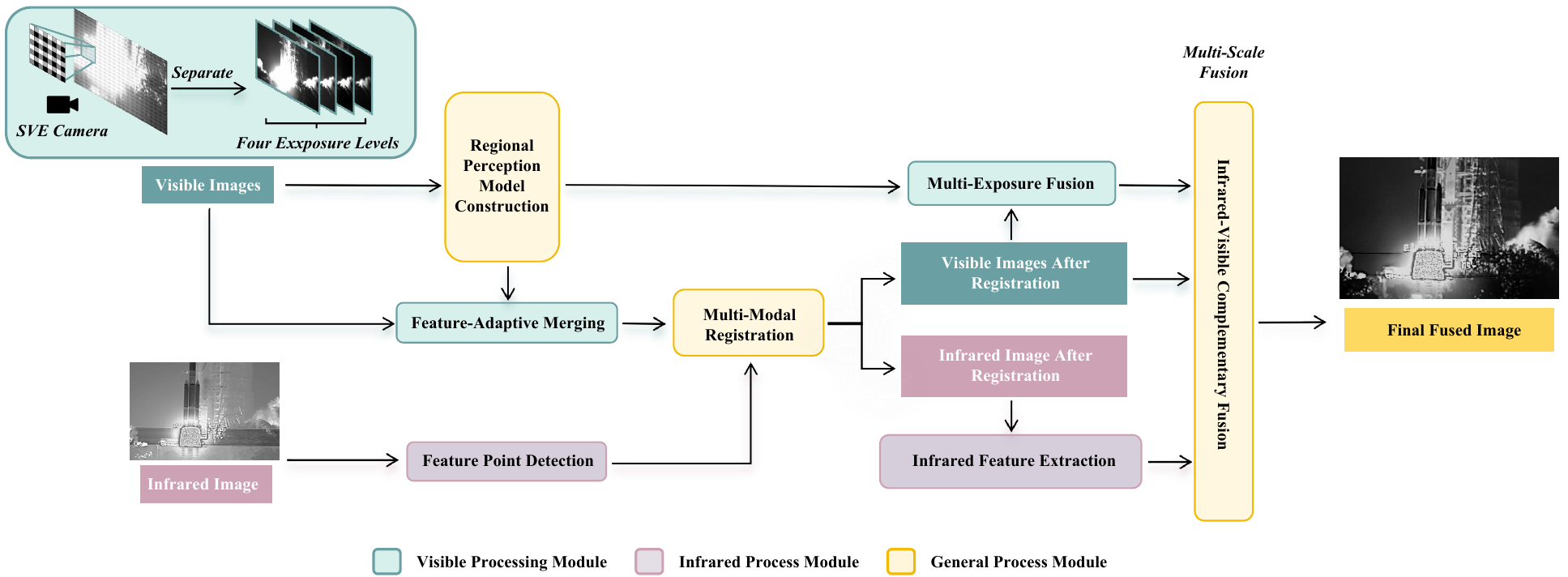}
	\caption{Overall framework for multi-modal complementary fusion in extreme scenarios.}
	\label{fig:MAIN}
\end{figure*}

The proposed framework integrates advanced imaging hardware to synergistically unify multi-exposure fusion and infrared-visible multi-modal fusion, addressing the critical demand for high-fidelity imaging in extreme scenarios. This framework comprises four main modules: regional perception model construction, feature adaptive merging, multi-exposure fusion, and infrared-visible complementary fusion. As shown in Fig. \ref{fig:taste}, our framework demonstrates superior fusion performance compared to state-of-the-art methods such as LDFusion \cite{LDFusion}. The results emphasize the effectiveness of the perceptual fusion strategy, which not only preserves key details but also reduces artifacts, even in challenging conditions where conventional deep learning-based methods struggle. The contributions of this paper are threefold:

\begin{itemize}
	\item[$\bullet$] A high dynamic range multi-spectral fusion framework with an SVE camera is proposed, enabling multi-exposure fusion and infrared compensation for high-quality imaging in challenging conditions.
	\item[$\bullet$] A feature-adaptive merging method uses brightness-dependent curve fitting to calculate dynamic merging weights from raw images, improving feature robustness and supporting multi-modal registration.
	\item[$\bullet$] A spatial SSIM-driven fusion method combines multi-exposure and visible-infrared data to compensate for missing information and enhance image quality.
\end{itemize}

The structure of this paper is as follows: Section II provides a detailed overview of the proposed framework. Section III explores the application of the framework in single-exposure scenarios. Section IV presents the experimental results, and Section V concludes the paper.

\section{Proposed Framework}
In this section, a comprehensive description of the proposed perceptual fusion framework is presented, as depicted in Fig. \ref{fig:MAIN}.

\subsection{Space-Variant Exposure Camera}
The space-variant exposure (SVE) camera technology was developed to address the increasing demand for high dynamic range (HDR) imaging in complex optical environments \cite{SVE}.
Specifically designed for the challenges of rocket launches, the SVE camera offers several advantages over traditional HDR systems \cite{2016Real,2016HDR,Static}. Its compact and portable design facilitates straightforward deployment, while its capability to capture images without parallax between sub-images at varying exposure levels obviates the need for alignment, thereby outperforming multi-camera composite configurations.

The primary innovation lies in the precisely engineered attenuator array positioned at the sensor plane, as illustrated in Fig. \ref{fig:MAIN}. Each macro-pixel comprises four micro-attenuators, which are configured with logarithmically scaled transmittance ratios (1:0.55:0.45:0.0025), thereby facilitating a 52 dB extension of the dynamic range. This spatially modulated attenuation allows for the concurrent capture of four distinct exposure levels using a single CMOS sensor array. The grid-aligned micro-attenuators maintain strict pixel-level correspondence, thereby ensuring photometric consistency across all exposure channels. As a result, the SVE camera is capable of capturing images at four different exposure levels in a single frame, thus enabling the acquisition of image sequences under varying exposure conditions, such as those encountered in hazy environments.

\subsection{Regional Perception Model Construction}
\begin{figure}[tp]
	\centering
	\includegraphics[width=0.48\textwidth]{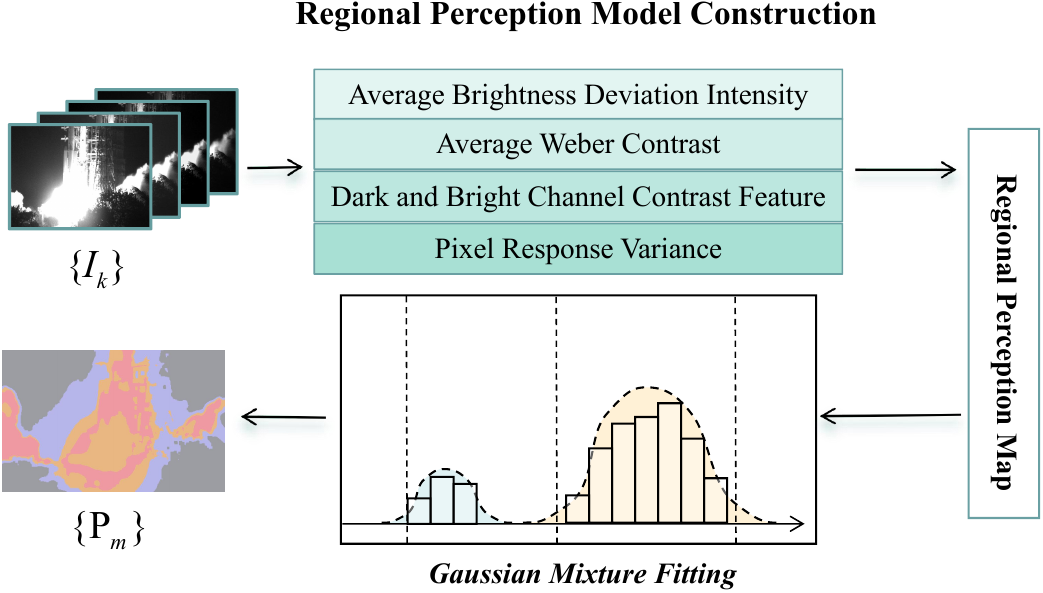}
	\caption{Flowchart of the regional perception model construction module, where different colors in the regional perception map represent distinct regions.}
	\label{fig:map}
\end{figure}

In extreme imaging scenarios, traditional methods often fail to meet the required standards. This limitation highlights the need for an integrated approach that utilizes local features for adaptive enhancement. Building on previous work \cite{2019Scene}, we propose a regional perception model that combines local and global features, incorporating gray-level, contrast, and variance information tailored to the SVE imaging device. This results in a four-dimensional feature perception model aimed at improving adaptability and robustness in image processing. The flowchart for this section is presented in Fig. \ref{fig:map}.

The model begins with dynamic range characterization using the average brightness deviation intensity ($BI$):
\begin{equation}
	BI(x) = \frac{1}{K}\sqrt {\sum\limits_{k = 1}^K {(max(} {{\rm{I}}_k}(x),T) - {\mu _k}{)^2}} 
	\label{eq:BI}
\end{equation}
where $k$ represents the image index in the sequence, ${{\rm{I}}_k}\left( x \right)$ denotes the gray value at pixel $x$ for the $k$-th image, and ${{\mu _k}}$ represents the mean intensity. The threshold $T = {\mu _k}/2$ is applied to filter out dark regions.

Local contrast preservation is quantified using the average Weber contrast ($WC$), which evaluates the sharpness of edges across multiple exposures:

\begin{equation}
	WC(x) = \frac{1}{K}\sum\limits_{k = 1}^K {\frac{{|\nabla {{\rm{I}}_k}(x)|}}{{{{\rm{I}}_k}(x) + 1}}}
	\label{eq:WC}
\end{equation}
where $\nabla $ denotes the gradient operator.

To mitigate environmental interference (e.g., uneven haze or combustion glare), we introduce a generalized contrast feature ($CF$) inspired by multi-exposure channel analysis. Unlike conventional dark/bright channel priors \cite{DCP, Wang2013, 2017}, our $CF$ extends these priors to multi-exposure sequences and avoids the need for the inverse derivation of the physical scattering model \cite{Optic, Non-local} for image restoration. The model is represented as:
\begin{equation}
	{\rm{I}}(x) = {\rm{J}}(x)t(x) + {\rm{A}}(1 - t(x))
	\label{eq:Model}
\end{equation}
where ${\rm{J}}(x)$ denotes the haze-free image, $t(x)$ is the transmission, and $\rm{A}$ is the atmospheric light. Assuming constant transmission and atmospheric light within a local block, the dark and bright channels are defined as:
\begin{equation}
	\left\{ \begin{array}{l}
		{{\rm{I}}^{\rm{d}}}(x) = t \cdot {{\rm{J}}^{\rm{d}}}(x) + {\rm{A}}(1 - t)\\
		{{\rm{I}}^{\rm{b}}}(x) = t \cdot {{\rm{J}}^{\rm{b}}}(x) + {\rm{A}}(1 - t)
	\end{array} \right.
	\label{eq:Jd_Jb}
\end{equation}
here, the dark and bright channels are derived by applying the minimum and maximum operators across the four exposure channels, respectively.
From this, we derive the $CF$ as:
\begin{equation}
	{CF(x) = 1 - \frac{{{{\rm{I}}^d}(x)}}{{\max ({{\rm{I}}^b}(x),1)}}}
	\label{eq:CF}
\end{equation}

Finally, the pixel response variance ($V$) is calculated using multi-exposure statistics, enabling the identification of regions with significant grayscale fluctuations in high dynamic range scenes. This approach effectively distinguishes fire regions and highlights reflections.
\begin{equation} 
	{V(x) = \frac{{{\rm{Var}}(x) - \bar \chi }}{{\bar \chi  + \varepsilon }}}
	\label{eq:V}
\end{equation}
where ${{\rm{Var}}(x)}$ is the pixel variance across different exposures, ${\bar \chi }$ is the geometric mean variance of the image, and ${\varepsilon }$ ensures numerical stability. This metric guides adaptive weight allocation for highlight suppression and detail recovery.

The final regional perception model is expressed as:
\begin{equation}
	{\rm{F}}(x) = \alpha BI(x) + \beta WC(x) + \gamma CF(x) + \sigma V(x)
	\label{eq:model}
\end{equation}
In this expression, $\alpha $, $\beta $, $\gamma $ and $\sigma$ represent the weighting factors, which are subject to the constraint $\alpha  + \beta  + \gamma  + \sigma  = 1$. In this model, all weights are set to 0.33.

The perception map is segmented in two stages: an initial segmentation based on the grayscale histogram, followed by Gaussian mixture fitting to refine the boundaries. To maintain consistency in the input and output sequence diagrams, the region is divided into four equal parts, with $M = 4$, thus ensuring a fixed structure for processing. This division results in a region set ${\rm{\{ }}{{\rm{P}}_m}{\rm{\} }}$ represents the index of each individual region.

\begin{figure}[tp]
	\centering
	\includegraphics[width=0.43\textwidth]{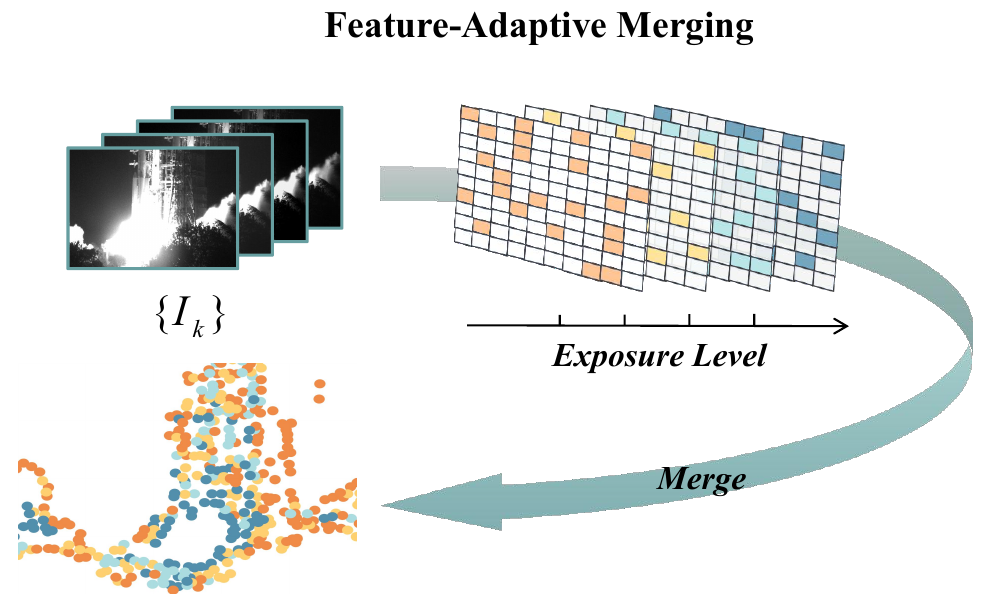}
	\caption{Features from images with different exposure levels are merged and filtered.}
	\label{fig:merge}
\end{figure}

\subsection{Feature Adaptive Merging}
Multi-modal imaging systems frequently encounter parallax-induced registration challenges in practical implementations. Although recent advancements in multi-modal registration \cite{Multimodal2015,Boosting,Unsupervised,MURF} have demonstrated promising results, feature extraction and fusion within spatially varying exposure (SVE) systems remain problematic. Our solution employs parallel feature extraction across multiple exposure levels followed by adaptive merging, preserving raw sensor data for descriptor computation to mitigate artifacts introduced by exposure fusion. Building upon \cite{Visual}, we enhance the framework through region-aware perception mapping, enabling effective management of illumination variations across exposure levels. A flowchart illustrating this process is provided in Fig. \ref{fig:merge}.

\begin{figure}[tp]
	\centering
	\includegraphics[width=0.43\textwidth]{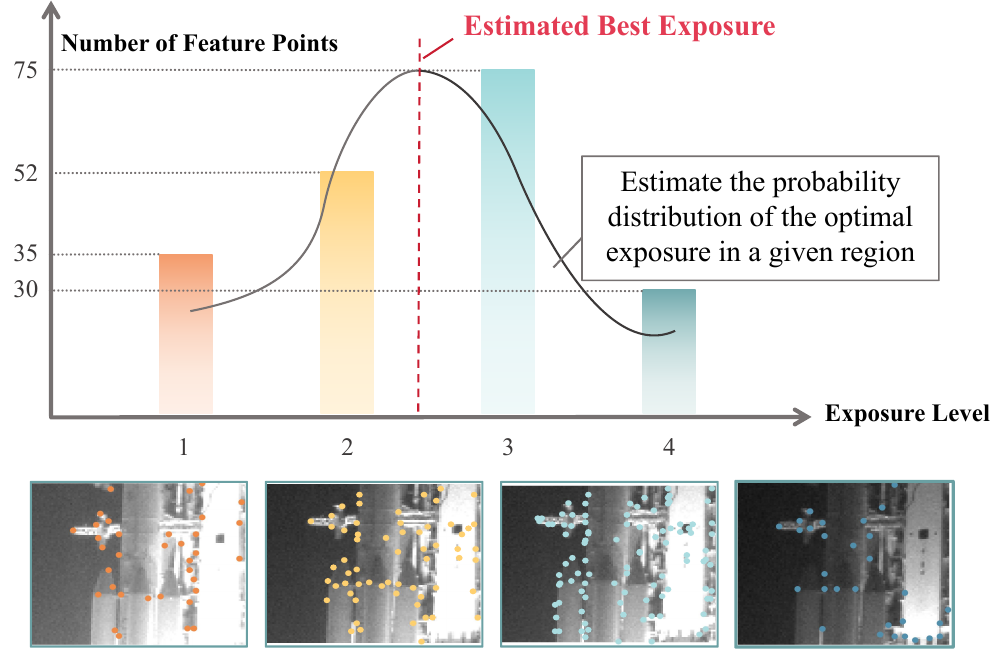}
	\caption{Estimate the probability distribution of the optimal exposure in a given region.}
	\label{fig:peak}
\end{figure}

The proposed approach is grounded in the Harris feature \cite{1988A, Heterologous}. Initially, an image pyramid is constructed, followed by the parallel extraction of Harris features from four images exhibiting different exposure levels. Feature points are then collected and assigned their corresponding exposure levels, which define the exposure for each feature. Unlike the method in \cite{Visual}, our regional perception mechanism enables independent processing of features within spatial partitions, significantly improving robustness in complex illumination scenarios.

Traditional maximum-response selection in multi-exposure fusion may result in instability across frames, especially when response values from different exposures are similar. To address this issue, we propose an adaptive filtering and merging strategy that leverages the exposure change curve of feature points. This strategy proves particularly effective in managing localized illumination variations through spatial-domain adaptation.

For the $m$-th sub-region, we count the number of feature points at each of the four exposure levels, denoted by $n_k^m$, where $k \in [1,4]$. The probability of feature distribution for the $m$-th sub-region is defined as:
\begin{equation}
	{L_m}(k) = \frac{{n_k^m}}{{\sum\nolimits_{j = 1}^4 {n_j^m} }}
\end{equation}
which represents the probability that a feature point in a specific region belongs to a particular exposure level.
The exposure transfer probability model ${E_m}(i{\left| j \right._{opt}})$ is given by:
\begin{equation}
	{E_m}(k\left| {{i_{opt}}} \right.) = \frac{1}{{\sqrt {2\pi } }}{e^{ - \frac{{{{(k - {i_{opt}})}^2}}}{{2{\sigma ^2}}}}}
\end{equation}
where ${{i_{opt}}}$ is the optimal exposure value for the region, determined by spline interpolation \cite{Some1985, Adaptive2017, Normal2009}. In this paper, the optimal exposure is obtained by fitting the brightness-feature point curve, as shown in Fig. \ref{fig:peak}.
The adaptive weight ${w_m}(i)$ is then calculated using the total probability formula:
\begin{equation} 
	{w_m}(k) = \sum\nolimits_{j = 1}^4 {{E_m}(k\left| {{i_{opt}}} \right.){L_m}(j)}
\end{equation}
The adaptive weight for region $m$ at different exposure levels $k$ can be obtained here. In this study, $K = M = 4$, resulting in a $4 \times 4$ adaptive weight matrix. This weight function ${w_m}(k)$ adjusts the response scores and selects the optimal feature points within each sub-region.  The fusion process incorporates two novel suppression criteria.

\begin{figure}[tp]
	\centering
	\includegraphics[width=0.48\textwidth]{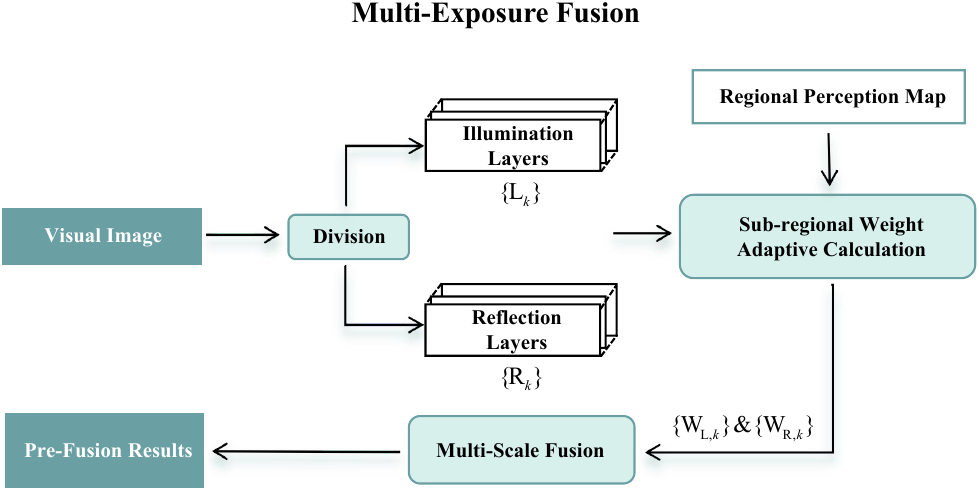}
	\caption{Multi-exposure fusion implementation idea flow chart.}
	\label{fig:MEF}
\end{figure}

\emph{Inter-frame maximum suppression}: At each spatial location across different exposure levels, the feature with the highest response value is retained. This approach mitigates the influence of illumination variations on the extraction of feature points, thereby enhancing the robustness of the process. Mathematically, for each location $(x,y)$, the following expression holds:
\begin{equation} 
	{f^*}(x,y) = \mathop {\arg \max }\limits_{k \in [1,4]} [{w_m}(k) \cdot S(x,y)]
\end{equation}
where ${w_m}(k)$ represents the weight value for the respective region and $S(x,y)$ is the original Harris response value.

\emph{Spatial maximum suppression}: After the initial fusion of points, we apply maximum suppression within a $3 \times 3$ region to prevent excessive point density in local areas. Only the points with the highest response value in each local region are retained. For each $3 \times 3$ region ${\Omega _{3 \times 3}}$,
\begin{equation} 
	{f^{**}}(x,y) = \mathop {\arg \max }\limits_{(x,y) \in {\Omega _{3 \times 3}}} [{f^*}(x,y)]
\end{equation}

After the selection of optimal features within each sub-region, these features are subsequently merged across all pyramid layers. To enhance the robustness of feature matching in multi-modal data, we integrate an improved multi-modal feature descriptor proposed by \cite{Heterologous}. This method employs an anisotropic weighted moment graph and the absolute phase congruency gradient histogram (HAPCG) to effectively address the challenges posed by significant illumination variations, contrast changes, and nonlinear radiation distortion commonly encountered in heterogeneous images. These procedures significantly contribute to the accurate registration of multi-exposure image sequences and infrared images, particularly in challenging and extreme imaging scenarios.

\subsection{Multi-Exposure Fusion}
Multi-exposure fusion (MEF) is essential for reconstructing high dynamic range (HDR) images in challenging illumination conditions, such as rocket launches, where haze and uneven brightness are common. This section presents an enhanced MEF technique that integrates a regional perception map to refine the fusion process.

The innovation lies in adaptive weight calculation and hierarchy-aware pyramid fusion. The algorithm flow is shown in Fig. \ref{fig:MEF}. First, the image is processed using Retinex \cite{Low-Light,Simultaneous} for division. Then, weights for specific regions in both the illumination and reflection layers are calculated to decouple haze suppression and detail enhancement. For the illumination layer, a dual-weighting mechanism combines gradient-based exposure quality (Eq. \ref{eq:w1}) with region-specific brightness constraints (Eq. \ref{eq:w2}), ensuring optimal exposure and haze suppression.

\begin{equation}
	{{\rm{W}}_{{{\rm{L}}_1},k}}(x) = \frac{{Gra{d_k}{{({{\rm{L}}_k}(x))}^{ - 1}}}}{{\sum\limits_{k = 1}^K G ra{d_k}{{({{\rm{L}}_k}(x))}^{ - 1}} + \varepsilon }}
	\label{eq:w1}
\end{equation}
\begin{equation}
	{{\rm{W}}_{{{\rm{L}}_2},k}}(x) = \frac{1}{\psi }\exp \left( { - \frac{{{{({{\rm{L}}_k}(x) - {u_k})}^2}}}{{2{\sigma ^2}}}} \right)
	\label{eq:w2}
\end{equation}
here, $Gra{d_k}({{\rm{L}}_k}(x))$ represents the gradient of the cumulative histogram for the illumination component ${{\rm{L}}_k}(x)$. ${\psi }$ is the normalized coefficient, ${{\sigma }}$ is the weight correction factor, and ${{u_k}}$ is adjusted based on the average brightness of each region. The term $\varepsilon$ is a small constant.

\begin{figure*}[tp]
	\centering
	\includegraphics[width=0.95\textwidth]{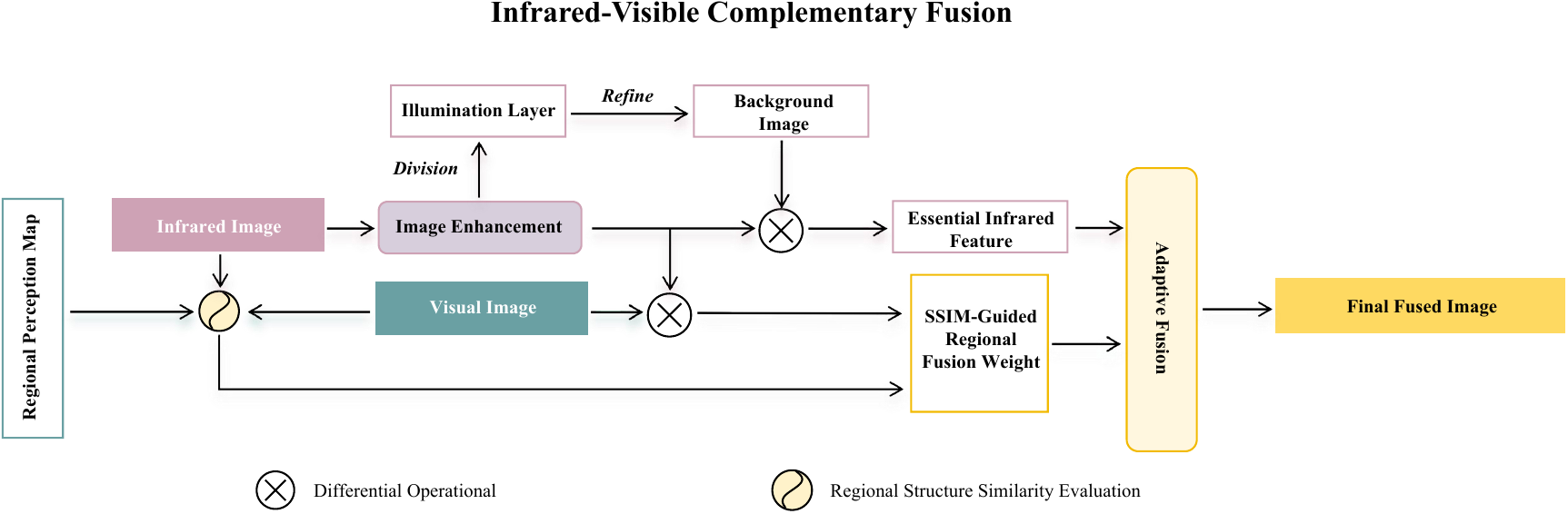}
	\caption{Flowchart of the infrared-visible complementary fusion concept.}
	\label{fig:IVFusion}
\end{figure*}

The computed weights are subsequently decomposed into Gaussian pyramids, while multi-scale image details are represented using Laplacian pyramids. The fusion process is applied simultaneously to both the brightness and detail layers, as well as the multi-scale image pyramids. At each scale, the fusion operation is formulated as follows:
\begin{small}
	\begin{equation}
		{R_{\rm{k}}}^{(l)} = \{ {L^{(l)}}\{ {{\rm{L}}_k}\}  \cdot {G^{(l)}}\{ {{\rm{W}}_{{\rm{L}},k}}\} \} \{ {L^{(l)}}\{ {{\rm{R}}_k}\}  \cdot {G^{(l)}}\{ {{\rm{W}}_{{\rm{R}},k}}\} \}  
	\end{equation}
\end{small}
In this equation, the final illumination weight $\{ {{\rm{W}}_{{\rm{L}},k}}\} $ is the product of ${{\rm{W}}_{{\rm{L}},k}} = {{\rm{W}}_{{{\rm{L}}_1},k}} \cdot {{\rm{W}}_{{{\rm{L}}_2},k}}$, and the detail layer weight $\{ {{\rm{W}}_{{\rm{R}},k}}\} $ is computed using improved guided filtering \cite{Simultaneous}. $G$ and $L$ denote the Gaussian and Laplacian pyramids, respectively, with $l$ representing the pyramid level.

The preliminary fusion result ${{\rm{I}}_{{\rm{pre}}}}$ integrates information from the HDR scene within the visible spectrum. The next step is to extend this method to multi-modal fusion, addressing a broader range of environmental complexities and enhancing robustness in extreme aerospace imaging scenarios.

\subsection{Infrared-Visible Complementary Fusion}
This section introduces a multi-modal fusion method based on a complementary mechanism, utilizing the aligned visible sequence and infrared images, along with the pre-fusion results of the visible images. As shown in Fig. \ref{fig:IVFusion}, the proposed algorithm integrates three key components: infrared feature extraction, SSIM-guided regional fusion weight calculation, and adaptive fusion. These components work together to produce the final fusion output.

\emph{1) Essential Infrared Feature Extraction:} Due to the unique characteristics of infrared imaging, target features are often distinguishable from the background based on brightness differences \cite{ZHANG2017,Thermal2023}. Therefore, extracting prominent targets from infrared images is crucial. To achieve this, we first reconstruct the infrared background and then separate the bright or dark features to extract the necessary infrared information.

Given the infrared image ${{\rm{I}}_{ir}}$, the objective is to preserve the structural information while smoothing the texture details, leading to the following optimization problem:
	\begin{equation}
		\mathop {\min }\limits_{{{\rm{B}}_{ir}}} \left\| {{{{\rm{\tilde I}}}_{ir}} - {{\rm{B}}_{ir}}} \right\|_2^2 + {c_1}\left\| {\nabla {{\rm{I}}_{ir}} - \nabla {{\rm{B}}_{ir}}} \right\|_2^2 + {c_2}{\left\| {{G_{ir}} \circ \nabla {{\rm{B}}_{ir}}} \right\|_1}
	\end{equation}
here ${{{{\rm{\tilde I}}}_{ir}}}$ is the image after preliminary smoothing via guided filtering \cite{WGIF}, $\nabla $ denotes the first-order derivative operator, and ${{G_{ir}}}$ is the weight matrix. ${c_1}$ and ${c_2}$ are regularization parameters. The weight matrix ${{G_{ir}}}$ is defined as:
\begin{equation}
	\begin{array}{l}
		{G_{ir,x}} =  - \log ({\nabla _x}{{{\rm{\tilde I}}}_{ir}})\\
		{G_{ir,y}} =  - \log ({\nabla _y}{{{\rm{\tilde I}}}_{ir}})
	\end{array} 
\end{equation}
here ${\nabla _x}$ and ${\nabla _y}$ are the horizontal and vertical gradients, respectively. The logarithmic function enhances the weight of texture details while reducing that of structural details, preserving the structure while smoothing the texture.
The optimization is solved using the alternating direction minimization (ADMM) algorithm \cite{Singh2020}. The smooth infrared background image (${{{\rm{B}}_{ir}}}$) is then obtained, and essential features are extracted by subtracting the background from the original infrared image:
\begin{equation}
	\zeta  = \max ({{\rm{I}}_{ir}} - {{\rm{B}}_{ir}},0)
	\label{eq:zeta}
\end{equation}

\emph{2) SSIM-Guided Regional Fusion Weight:}
In the previous section, prominent features were extracted from the infrared image. Here, we integrate both visible and infrared images to leverage the complementary features of each modality. Building on the work of \cite{IVFusion} and \cite{ZHANG2017}, we propose a method for extracting complementary information using a regional SSIM operator in conjunction with regional perception maps. The key innovation of this approach lies in the integration of structural similarity quantification and regional perception analysis, which culminates in a three-stage processing framework.

\begin{figure*}[tp]
	\centering
	\includegraphics[width=0.85\textwidth]{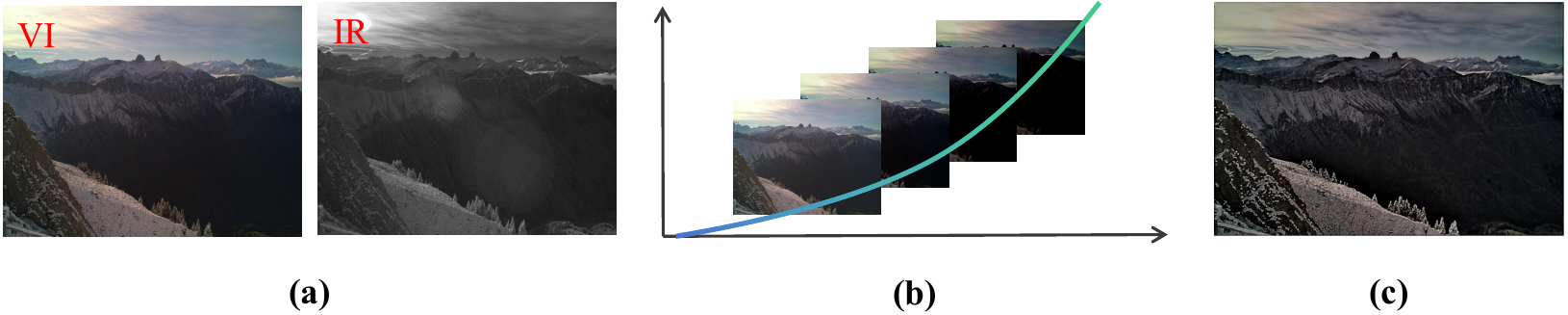}
	\caption{Schematic representation of single-exposure expansion method: (a) Input image; (b) Gamma-transformed intermediate; (c) Fusion result.}
	\label{fig:single}
\end{figure*}	
\begin{figure}[tp]
	\centering
	\includegraphics[width=0.4\textwidth]{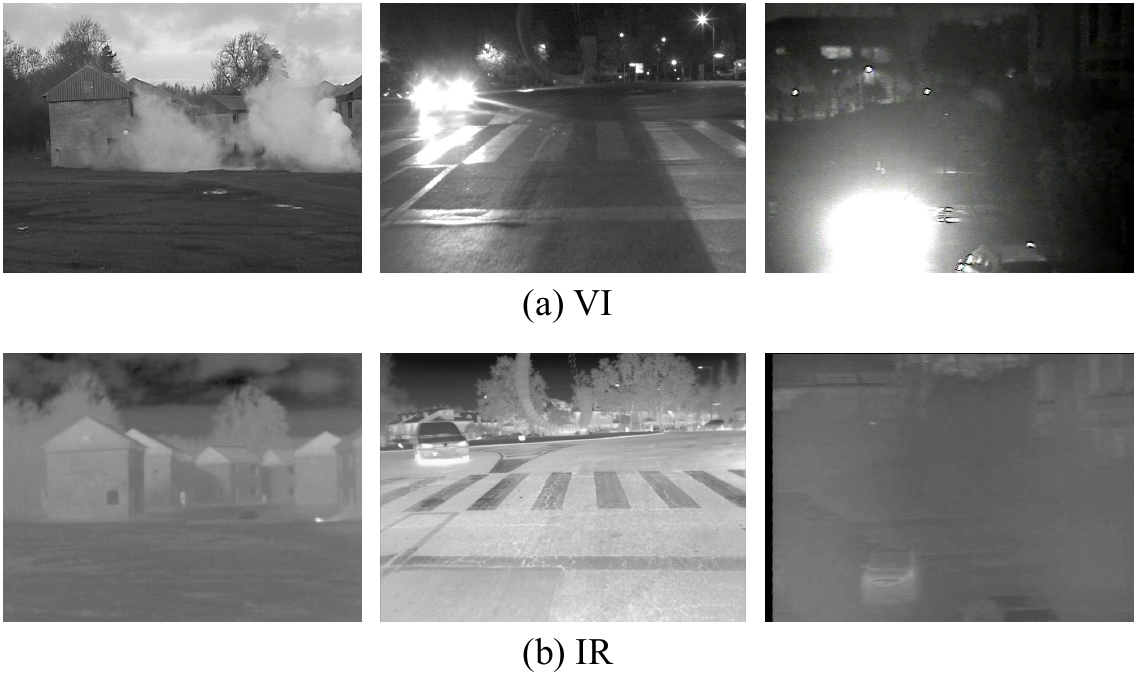}
	\caption{Selected data pairs from \textit{Public}.}
	\label{fig:dataset}
\end{figure}	

For each segmented region ${{\rm{P}}_m}$, we define a novel structural coherence metric using a spatial-constrained SSIM evaluation:
\begin{equation}
	Scor{e_m} = LS({{\rm{I}}_{pre}} \odot {{\rm{P}}_m},{{\rm{I}}_{ir}} \odot {{\rm{P}}_m},N,t)
	\label{eq:score}
\end{equation}
here, ${{\rm{I}}_{pre}} \odot {{\rm{P}}_m}$ represents the application of the mask ${{\rm{P}}_m}$ to extract region-specific pixels, and $LS( \cdot )$ denotes the local SSIM operator. A sliding window of size $N \times N$ is employed, moving by $t$ pixels in both horizontal and vertical directions, thereby partitioning the images into patches. This produces a score map that represents the local similarity between the visible and infrared images in region $m$. The SSIM between the corresponding image patches is calculated as follows:
	\begin{equation}
		{\rm{SSIM(}}i,j{\rm{)}} = \frac{{(2{u_i}{u_j} + {b_1})(2{\sigma _{ij}} + {b_2})}}{{({u_i}^2 + {u_j}^2 + {b_1})({\sigma _i}^2 + {\sigma _j}^2 + {b_2})}}
	\end{equation}
where ${\rm{(}}i,j{\rm{)}}$ refers to image pair (infrared vs. pre-fusion), ${{u_i}}$ and ${{u_j}}$ are the pixel averages of images $i$ and $j$, ${{\sigma _i}}$ and ${{\sigma _j}}$ are the standard deviations, ${{\sigma _{ij}}}$ is the covariance, and ${{b_1}}$ and ${{b_2}}$ are constants used to prevent division by zero. Smaller SSIM values indicate lower structural similarity, suggesting that the infrared image should dominate in such regions.
To enhance regions with high structural consistency (e.g., sharp edges) and suppress artifacts in areas with low correlation (e.g., haze-obscured zones), we apply a quadratic mapping:
\begin{equation}
	{w_m} = {(Scor{e_m})^2}
	\label{eq:wm}
\end{equation}

In this way, the structural similarity of the image can be evaluated finely, so as to achieve higher accuracy and better visual effect in image processing and fusion tasks.

\emph{3) Adaptive Fusion:}
After extracting essential infrared features and calculating SSIM-guided regional fusion weights, the final step involves integrating these components through an adaptive fusion strategy. This section details the adaptive fusion process, which combines the extracted infrared features with the visible image sequence, guided by the calculated regional weights to produce the final fusion result.

By combining Eq. (\ref{eq:zeta}) and Eq. (\ref{eq:wm}), we extract complementary information from infrared and visible images, with weights applied to adjust the credibility of the information.
\begin{equation}
	{{\zeta '}_k} = \zeta  + {w_m} \cdot \max ({{\rm{I}}_k} - {{\rm{I}}_{ir}},0)
\end{equation}
here ${{\rm{I}}_k}$ denotes the $i$-th visible image in the sequence. The term max $\max ({{\rm{I}}_k} - {{\rm{I}}_{ir}},0)$ represents the difference between the infrared and visible images, which is subsequently scaled by the weight matrix ${w_m}$ to emphasize regions exhibiting high structural similarity while mitigating artifacts.

The integrated features ${\zeta _{fused}}$ from each visible image are combined using a maximum fusion strategy to retain the most prominent features across all sub-images:
\begin{equation}
	{\zeta _{fused}} = \max ({{\zeta '}_1},{{\zeta '}_2}, \cdots {{\zeta '}_K})
\end{equation}
This step ensures that the final fusion result retains the most salient features from each modality, enhancing the overall quality and information content of the fused image.
	
To address the low quality of infrared image data, we apply guided filtering to enhance result robustness. Additionally, we optimize the final fusion by performing brightness mapping on complementary features:
\begin{equation}
	{{\rm{I}}_{out}} = {{\rm{I}}_{pre}} + \eta  \cdot GIF({\zeta _{fused}})
\end{equation}
here $GIF( \cdot )$ is the guided filtering operator. In this paper, we use the improved guided filtering in \cite{Simultaneous}. ${{\rm{I}}_{out}}$ is final fused image. $\eta $ denotes the feature adjustment ratio, which is calculated as:
\begin{equation}
	\eta  = \min (\frac{{255}}{{Ave{r_{0.5}}}},1)
\end{equation}
here $Ave{r_{0.5}}$ represents the average brightness of the first 50 \% of the brightness values from the visible image ${{\rm{I}}_{pre}}$ and the fusion feature ${\zeta _{fused}}$, which ensures that the fused image maintains a balanced dynamic range.

\begin{figure*}[tp]
	\centering
	\includegraphics[width=0.98\textwidth]{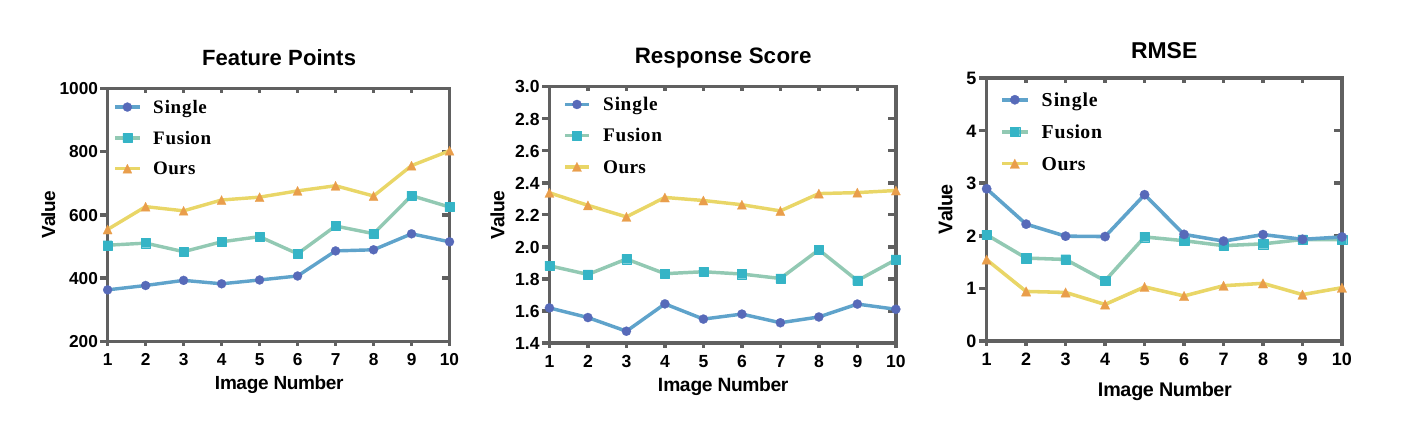}
	\vspace{-0.5cm}
	\caption{A quantitative comparison of our feature merging method with single-exposure and HDR fusion sequences for feature point detection. A higher number of feature points and response scores indicate better performance, while a smaller RMSE indicates better accuracy.}
	\label{fig:merge-1}
	\vspace{0.35cm}
		\centering
		\includegraphics[width=0.9\textwidth]{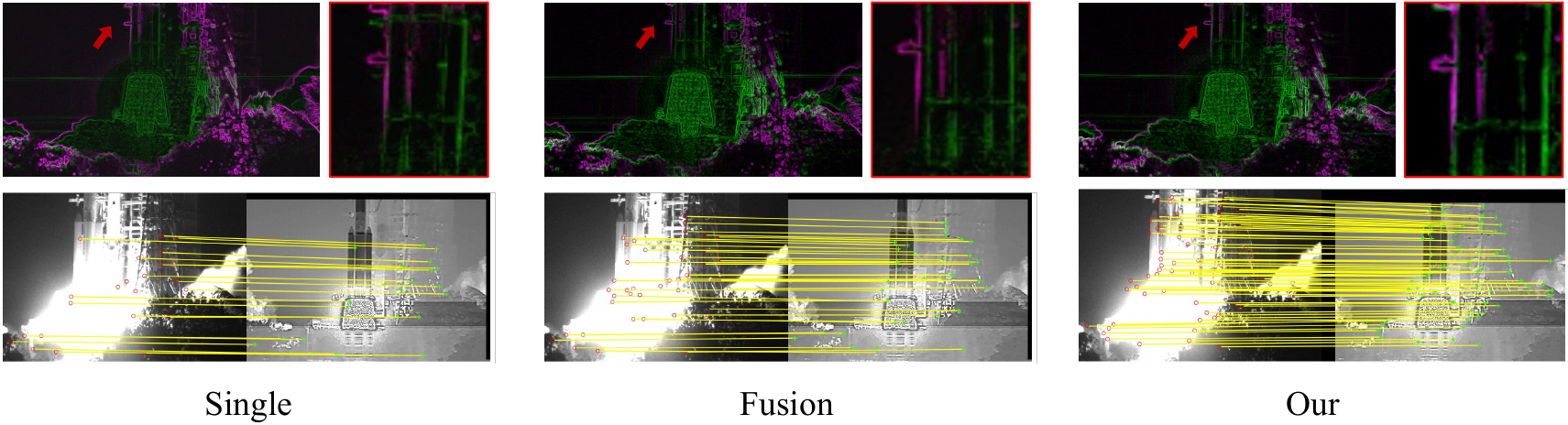}
		\caption{Qualitative registration results for multi-modal image pairs. First row: gradient overlay image of the image pair (magenta represents the visible gradient image, and the green represents the infrared gradient image). Second row: results of visible-infrared multi-modal registration.}
		\label{fig:merge-2}
	\end{figure*}

\section{Extended Applications in Single-Exposure Scenarios}
The SVE camera system demonstrates operational advantages in multi-exposure image acquisition through its parallel capture mechanism. However, this architecture entails inherent limitations: (1) inherent color information degradation constraining acquisition to grayscale modalities; (2) restricted generalizability due to its specialized hardware implementation. To overcome these constraints, we propose a generalized framework extending the methodology to conventional single-exposure datasets, thereby enhancing cross-platform compatibility and operational scope across diverse imaging conditions.

The proposed method for single-exposure expansion is based on the gamma transformation framework from \cite{GALDRAN2018}. Each input frame undergoes gamma correction to simulate pseudo-multi-exposure sequences using a parametric power-law operation:
\begin{equation}
	{\rm{I}}(x) \to \alpha  \cdot {\rm{I}}{(x)^r}
\end{equation}
where $\alpha $, $\gamma $ are positive real-valued coefficients. Varying $\gamma $ simulates diverse exposure conditions: $\gamma  > 1$ reduces high-luminance areas, while $\gamma  < 1$ enhances low-intensity regions. This nonlinear mapping preserves edge information and adapts local contrast characteristics.

For the color image shown in Fig. \ref{fig:single}, a color space transformation is first applied to extract the brightness component. This is accomplished through the following modifications:
\begin{equation}
	Y = 0.299R + 0.587G + 0.114B
\end{equation}
where $R$, $G$, and $B$ are the normalized RGB chromatic components. In high dynamic range scenarios affected by haze interference—often resulting in luminance saturation—the framework applies selective gamma correction ($\gamma  > 1$) to suppress highlights. This investigation employs an innovative adaptive parameter selection mechanism, establishing $\gamma $ values within [0.8, 1.4] through context-aware optimization, thereby outperforming conventional fixed-parameter approaches \cite{Simultaneous,SCENS}.

The multi-exposure sequence is processed using a multi-scale fusion technique that integrates the optimal regions from each exposure level into a unified output. This method effectively preserves color information while enhancing visibility in areas affected by haze or low contrast. Experimental results demonstrate the effectiveness of the approach, achieving infrared compensation and significantly improving overall image quality. Extending it to single-exposure datasets overcomes the limitations of the SVE camera, retaining the advantages of multi-exposure fusion and preserving color details.

\begin{figure*}[tp]
	\centering
	\includegraphics[width=0.9\textwidth]{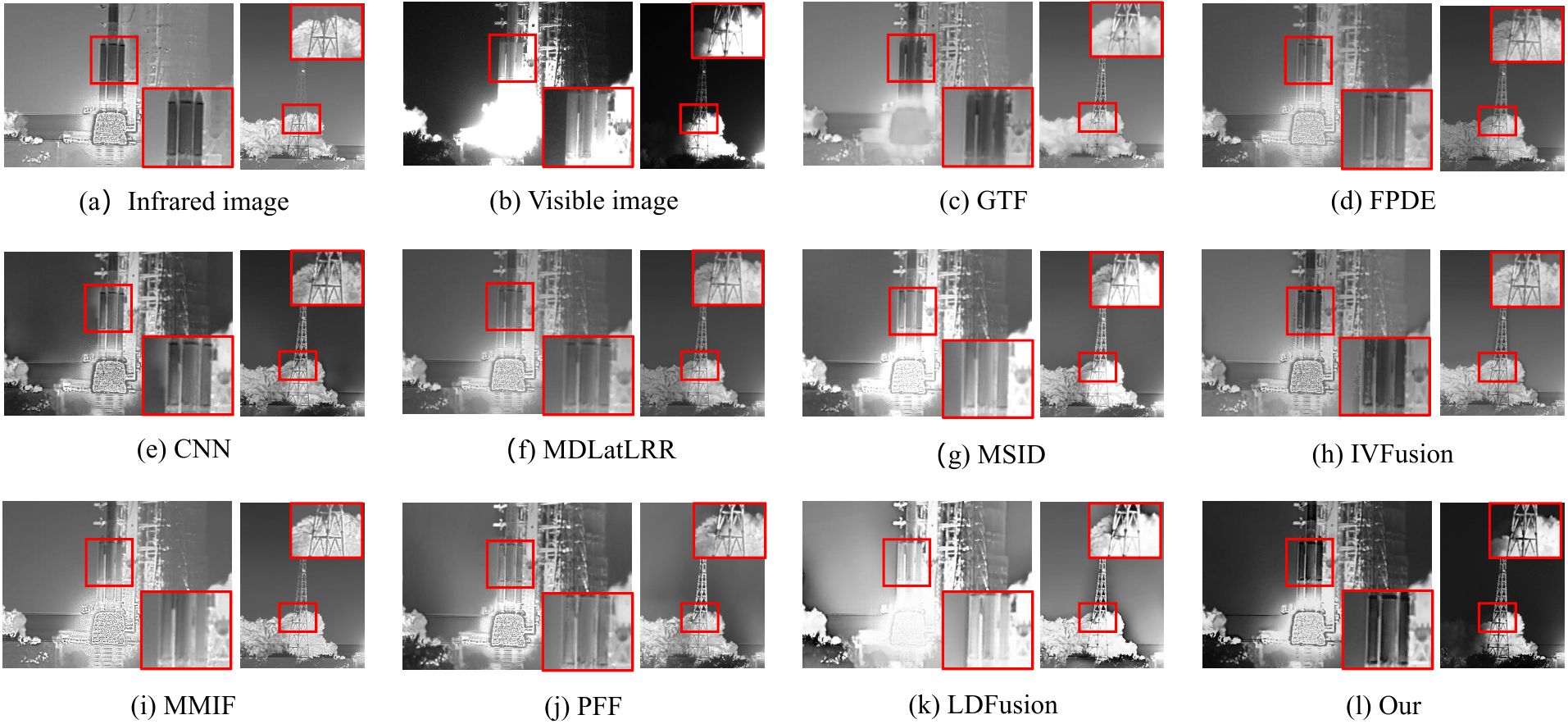}
	\caption{Qualitative comparisons of different fusion methods and our method on \textit{Rocket-1} image pairs.}
	\label{fig:fusion-1}
	%
\end{figure*}

 \section{Experiments}
	This section outlines the evaluation protocol used to validate the proposed framework. We begin with a description of the experimental setup and datasets. The effectiveness of the feature merging mechanism for multi-modal registration is then assessed. We proceed with both qualitative and quantitative comparisons against advanced algorithms. Finally, we highlight the framework's advantages in preserving image features and discuss its potential applications in other domains.	
	
	\subsection{Datasets and settings}
	Due to the specialized nature of the SVE camera-based methodology, conventional benchmark datasets are not suitable for validation. Therefore, an image sequence was acquired using the SVE imaging system, serving as the validated dataset for this study. Furthermore, to evaluate the extended algorithm described in Section III, self-collected rocket data, along with scene-like data from publicly available datasets, are utilized.
	
	\emph{1) Rocket-1:} A multispectral dataset from an SVE camera and a long-wave infrared camera at a launch site, consisting of 812 images from a nighttime rocket launch, each with a resolution of 2448 $\times $ 2048 pixels per frame.
	
	\emph{2) Rocket-2:}  Compared to \textit{Rocket-1}, this dataset consists of launch data from the same scene captured at different time periods. It contains a total of 560 daytime rocket launch images, each with a resolution of 1224$\times $1024.
	
	\emph{3) Public:} A test dataset of 60 samples selected from public datasets (TNO \cite{TNO}, LLVIP \cite{LLVIP}, RoadScene \cite{U2Fusion}) under extreme brightness and haze conditions, as shown in Fig. \ref{fig:dataset}.
	
	We compare our framework with nine state-of-the-art fusion methods: GTF \cite{GTF}, FPDE \cite{FPDE}, CNN \cite{CNN}, MDLatLRR \cite{MDLatLRR}, MSID \cite{MSID}, IVFusion \cite{IVFusion}, MMIF \cite{MMIF}, PFF \cite{PFF}, and LDFusion \cite{LDFusion}. Fusion performance is evaluated using six metrics: average gradient (AG) \cite{CUI2015199}, mutual information (MI) \cite{MI}, peak signal-to-noise ratio (PSNR) \cite{PNSR}, visual information fidelity for fusion (VIFF) \cite{VIFF}, natural image quality evaluator (NIQE) \cite{NIQE}, and multi-exposure fusion structural similarity index (MEF-SSIM) \cite{mef-ssim}. AG measures detail saliency, MI quantifies information transfer, PSNR evaluates distortion, VIFF, and NIQE assess visual fidelity, and MEF-SSIM evaluates structural similarity for multi-exposure fusion.
	
	\subsection{Feature Merging Validation}
	This section presents a comparative analysis of our proposed feature merging method against two alternatives: the single exposure sequence and the high dynamic range (HDR) fusion sequence, used as inputs for feature point detection. For the HDR fusion method, we apply Mertens' approach \cite{Mertens} to synthesize HDR images and incorporate them into the HDR fused sequences. Feature detection robustness is assessed using three metrics: feature point number, mean response magnitude, and root mean square error (RMSE).
	
	As depicted in Fig. \ref{fig:merge-1}, our feature merging approach demonstrates superior performance across all evaluation metrics, generating more robust and accurate feature points. In contrast to single-exposure imaging, HDR fusion markedly enhances the stability of feature extraction and highlights the influence of illumination variations on descriptor reliability. Nevertheless, current HDR implementations lack additional filtering mechanisms to ensure the robustness of feature points against photometric distortions. Our proposed methodology comprehensively addresses these interfering variables through systematic parameter optimization, achieving superior outcomes in challenging illumination environments. These results underscore the benefits of multi-exposure fusion, particularly in scenes with extreme lighting conditions.
	
	To further validate our method, registration experiments are conducted on features from both single-exposure and HDR fusion images. As illustrated in Fig. \ref{fig:merge-2}, gradient superposition analysis reveals misalignment in the rocket contour during single exposure registration, and HDR fusion also suffers from misregistration. By contrast, our method significantly enhances registration accuracy by utilizing a greater number of robust feature points, stronger constraints, and improved precision. This approach not only resolves the misalignment issues but also ensures a more accurate and reliable fusion process, thereby demonstrating the innovation and professionalism of our feature adaptive merging methodology.

\begin{figure*}[tp]
	\centering
	\includegraphics[width=0.9\textwidth]{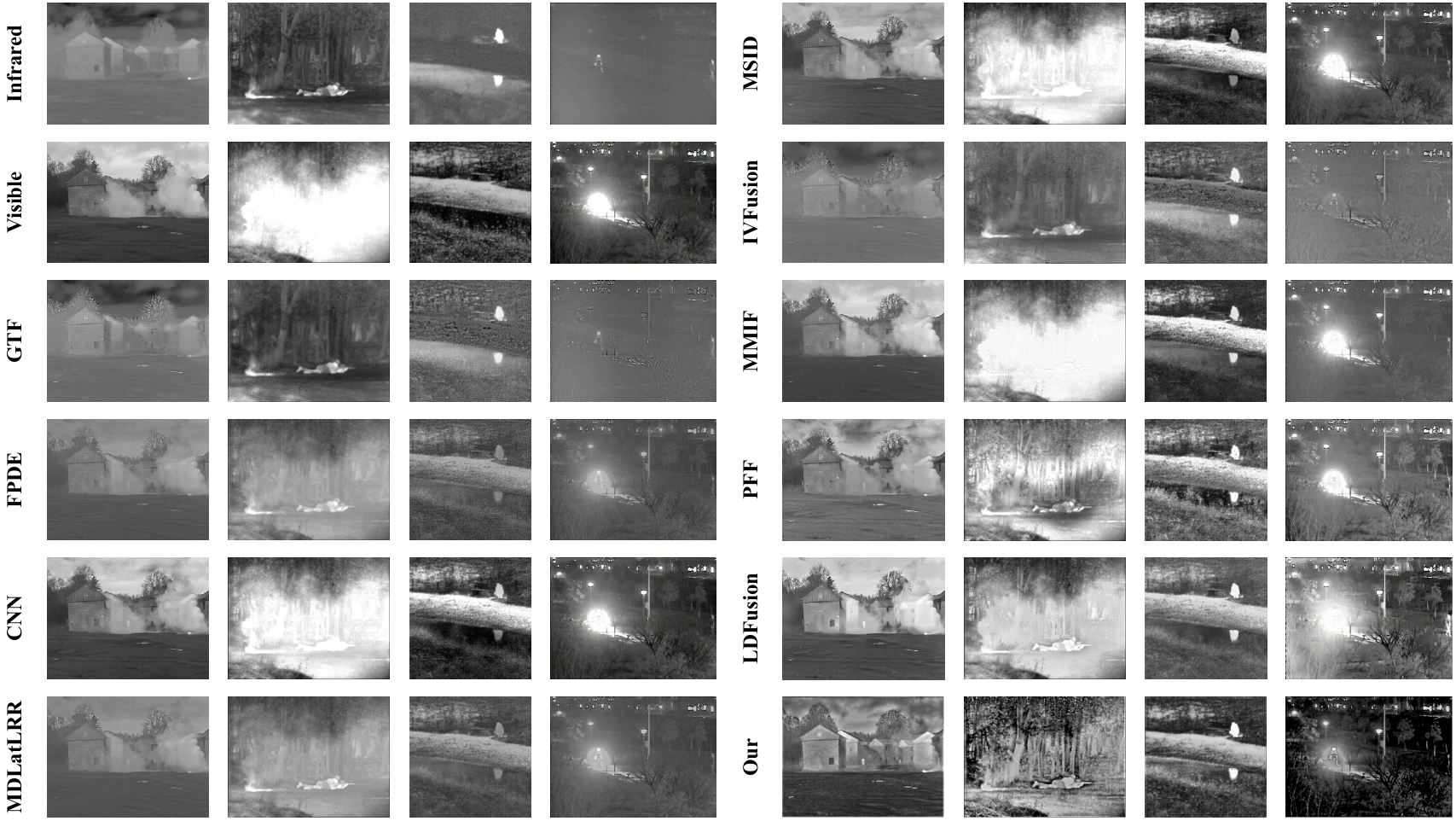}
	\caption{Qualitative comparisons of different fusion methods and our method on \textit{Public} image pairs.}
	\label{fig:fusion-3}
\end{figure*}
		
	\subsection{Qualitative Comparison}
	Fig. \ref{fig:fusion-1} provides a qualitative comparative analysis between the proposed methodology and eight representative techniques applied to launch scene image pairs. Our framework demonstrates remarkable superiority. As depicted in Fig. \ref{fig:fusion-1}, the multi - exposure fusion enhancement and fidelity preservation mechanisms in our framework not only significantly improve target visibility, but also retain fine details, such as the structural features of the signal towers. Moreover, the use of infrared data compensates for information loss in overexposed regions, thus ensuring the completeness of the image. The comparative analysis indicates that our algorithm exhibits significantly fewer fusion artifacts than the comparison methods, with this performance improvement attributable to the adaptive brightness equalization module and gradient-domain enhancement architecture embedded in the proposed framework.
	
	Fig. \ref{fig:fusion-3} presents additional fusion results from the \textit{Public} datasets. The results reveal that MMIF, LDFusion, and CNN have limitations in integrating information in HDR scenes, negatively impacting target detection (e.g., the individual in the fourth image set), suggesting the constrained generalization of deep learning-based fusion methods. GTF and IVFusion fail to adequately integrate visible information, as seen in the incomplete fusion of background details in the fourth image set. While FPDE and MDLatLRR integrate some information, their pixel-level fusion causes gradient information loss and overall image degradation. MSID preserves targets in hazy environments but struggles with saturated regions (e.g., the individual in the fourth column). In contrast, the proposed method consistently outperforms these approaches, maintaining target integrity in hazy scenes while effectively suppressing haze. In HDR scenes, it reduces saturation, enhances visible details, and compensates for infrared data, thereby ensuring superior structural integrity and more detailed target information.

	\begin{table*}[tp]
	\centering
	\caption{Quantitative assessment using different metrics, with the best results highlighted in red and the second best in blue.}
	\vspace{-0.8cm}
	\renewcommand\arraystretch{1.2} 
	\centering   
	\resizebox{0.85\textwidth}{!}{ 
		\begin{tabular}{cccccccccccc}
			&       &       &       &       &       &       &       &       &       &       &  \\
			&       &       &       &       &       &       &       &       &       &       &  \\
			\midrule
			\textbf{Data} & \textbf{Metric} & \textbf{GTF} & \textbf{FPDE} & \textbf{CNN} & \textbf{MDLatLRR} & \textbf{MSID} & \textbf{IVFusion} & \textbf{MMIF} & \textbf{PFF} & \textbf{LDFusion} & \textbf{Our} \\
			\midrule
			\multirow{6}[2]{*}{\textit{Rocket-1}} & \textbf{AG$\uparrow$} & 1.129  & 1.730  & 2.388  & 1.663  & 2.191  & 1.883  & 2.373  & 2.455  & \textcolor[rgb]{ 1,  0,  0}{\textbf{2.777 }} & \textcolor[rgb]{ 0,  .439,  .753}{\textbf{2.584 }} \\
			& \textbf{MI$\uparrow$} & 3.626  & 2.790  & 2.500  & 2.863  & 3.165  & \textcolor[rgb]{ 0,  .439,  .753}{\textbf{3.747 }} & 3.789  & 1.800  & 2.013  & \textcolor[rgb]{ 1,  0,  0}{\textbf{4.155 }} \\
			& \textbf{PNSR$\uparrow$} & 63.625  & \textcolor[rgb]{ 1,  0,  0}{\textbf{64.269 }} & 63.717  & 64.380  & 63.665  & 63.975  & 63.801  & 61.255  & 60.142  & \textcolor[rgb]{ 0,  .439,  .753}{\textbf{63.887 }} \\
			& \textbf{VIFF$\uparrow$} & 0.340  & 0.443  & 0.761  & 0.516  & \textcolor[rgb]{ 1,  0,  0}{\textbf{0.773 }} & 0.534  & 0.647  & 0.709  & 0.750  & \textcolor[rgb]{ 0,  .439,  .753}{\textbf{0.765 }} \\
			& \textbf{NIQE$\downarrow$} & 4.993  & 5.326  & 5.773  & 5.212  & 5.332  & 5.737  & 5.060  & 5.279  & \textcolor[rgb]{ 0,  .439,  .753}{\textbf{4.144 }} & \textcolor[rgb]{ 1,  0,  0}{\textbf{4.113 }} \\
			& \textbf{MEF-SSIM$\uparrow$} & 0.934  & 0.962  & \textcolor[rgb]{ 1,  0,  0}{\textbf{0.986 }} & 0.972  & 0.973  & 0.975  & \textcolor[rgb]{0,  .439,  .753}{\textbf{0.974 }} & 0.960  & 0.795  & 0.969  \\
			\midrule
			\multirow{6}[2]{*}{\textit{Rocket-2}} & \textbf{AG$\uparrow$} & 0.012  & 0.536  & 3.664  & 2.494  & 3.618  & 2.354  & 2.723  & 3.182  & \textcolor[rgb]{ 0,  .439,  .753}{\textbf{3.747 }} & \textcolor[rgb]{ 1,  0,  0}{\textbf{5.581 }} \\
			& \textbf{MI$\uparrow$} & 0.088  & 1.112  & \textcolor[rgb]{ 1,  0,  0}{\textbf{2.714 }} & 2.231  & 1.673  & 0.932  & 1.988  & 1.625  & \textcolor[rgb]{ 0,  .439,  .753}{\textbf{2.303 }} & 2.265  \\
			& \textbf{PNSR$\uparrow$} & 56.734  & 55.903  & 56.243  & 55.514  & 55.213  & 54.837  & 54.408  & 53.954  & \textcolor[rgb]{ 0,  .439,  .753}{\textbf{56.735 }} & \textcolor[rgb]{ 1,  0,  0}{\textbf{57.463 }} \\
			& \textbf{VIFF$\uparrow$} & 0.274  & 0.271  & 0.770  & 0.707  & 1.053  & 0.472  & 0.863  & 0.733  & \textcolor[rgb]{ 0,  .439,  .753}{\textbf{0.865 }} & \textcolor[rgb]{ 1,  0,  0}{\textbf{1.492 }} \\
			& \textbf{NIQE$\downarrow$} & 6.402  & 6.445  & 4.443  & 4.462  & 4.945  & 4.664  & 5.494  & 4.373  & \textcolor[rgb]{ 0,  .439,  .753}{\textbf{4.150 }} & \textcolor[rgb]{ 1,  0,  0}{\textbf{3.589 }} \\
			& \textbf{MEF-SSIM$\uparrow$} & 0.640  & 0.636  & \textcolor[rgb]{ 1,  0,  0}{\textbf{0.958 }} & 0.919  & 0.937  & 0.830  & 0.893  & 0.877  & 0.640  & \textcolor[rgb]{ 0,  .439,  .753}{\textbf{0.930 }} \\
			\midrule
			\multirow{6}[1]{*}{\textit{Public}} & \textbf{AG$\uparrow$} & 4.196  & 4.474  & 4.777  & 3.068  & 5.081  & 3.651  & 5.075  & \textcolor[rgb]{ 0,  .439,  .753}{\textbf{5.444 }} & 5.245  & \textcolor[rgb]{ 1,  0,  0}{\textbf{5.465 }} \\
			& \textbf{MI$\uparrow$} & 1.195  & 1.691  & \textcolor[rgb]{ 1,  0,  0}{\textbf{2.577 }} & 1.843  & 1.962  & 1.059  & 2.128  & 1.357  & 1.620  & \textcolor[rgb]{ 0,  .439,  .753}{\textbf{2.442 }} \\
			& \textbf{PNSR$\uparrow$} & 57.625  & \textcolor[rgb]{ 1,  0,  0}{\textbf{58.071 }} & 57.684  & 57.729  & 57.642  & 56.923  & 57.413  & 56.790  & 55.133  & \textcolor[rgb]{0,  .439,  .753}{\textbf{57.757 }} \\
			& \textbf{VIFF$\uparrow$} & 0.214  & 0.236  & 0.390  & 0.515  & 0.719  & 0.434  & 0.488  & \textcolor[rgb]{ 0,  .439,  .753}{\textbf{0.991 }} & 0.619  & \textcolor[rgb]{ 1,  0,  0}{\textbf{1.266 }} \\
			& \textbf{NIQE$\downarrow$} & 6.966  & 6.071  & 6.018  & 5.603  & 4.770  & 5.400  & 5.341  & 5.869  & \textcolor[rgb]{ 0,  .439,  .753}{\textbf{4.358 }} & \textcolor[rgb]{ 1,  0,  0}{\textbf{4.045 }} \\
			& \textbf{MEF-SSIM$\uparrow$} & 0.780  & 0.868  & 0.931  & 0.906  & 0.945  & 0.844  & \textcolor[rgb]{ 1,  0,  0}{\textbf{0.909 }} & 0.872  & 0.529  & \textcolor[rgb]{ 0,  .439,  .753}{\textbf{0.908 }} \\
			\bottomrule 
	\end{tabular}}
	\begin{tablenotes}            
		\item The symbol $\uparrow$ indicates that higher metric values are better, while $\downarrow$ indicates that lower values are better.
	\end{tablenotes}   
	\label{tab:table1}%
\end{table*}

	\subsection{Quantitative Comparison}
	This section selects 20 image pairs from each dataset for objective evaluation. The average results for each metric are demonstrated in Table \ref{tab:table1}, where optimal values are denoted in bold red and suboptimal counterparts in deep blue.
	
    The proposed framework consistently outperforms other methods across most evaluation metrics (see Table \ref{tab:table1}), particularly in the NIQE metric, which underscores its capability to enhance image quality while effectively minimizing information entropy loss. Furthermore, our method attains superior results in the AG, PSNR, and VIFF metrics, demonstrating enhanced visual contrast and greater detail in the fused images. The observed reduction in MI for the \textit{Rocket-2} dataset can be attributed to the adaptive infrared saliency filtering, which isolates thermal features while minimizing noise. Similarly, the decrease in MEF-SSIM for the \textit{Rocket-1} dataset is a result of the limited utilization of infrared background information in the nighttime scene, which consequently leads to a lower structural similarity relative to the original infrared image.
	In conclusion, the proposed framework generates images with superior visual quality, enhanced informational content, and better preservation of VIS image details, supporting subsequent measurement tasks. It also offers improved target contrast, robustness, and effective adaptation to challenging conditions.
	
	\begin{figure*}[tp]
		\centering
		\includegraphics[width=0.86\textwidth]{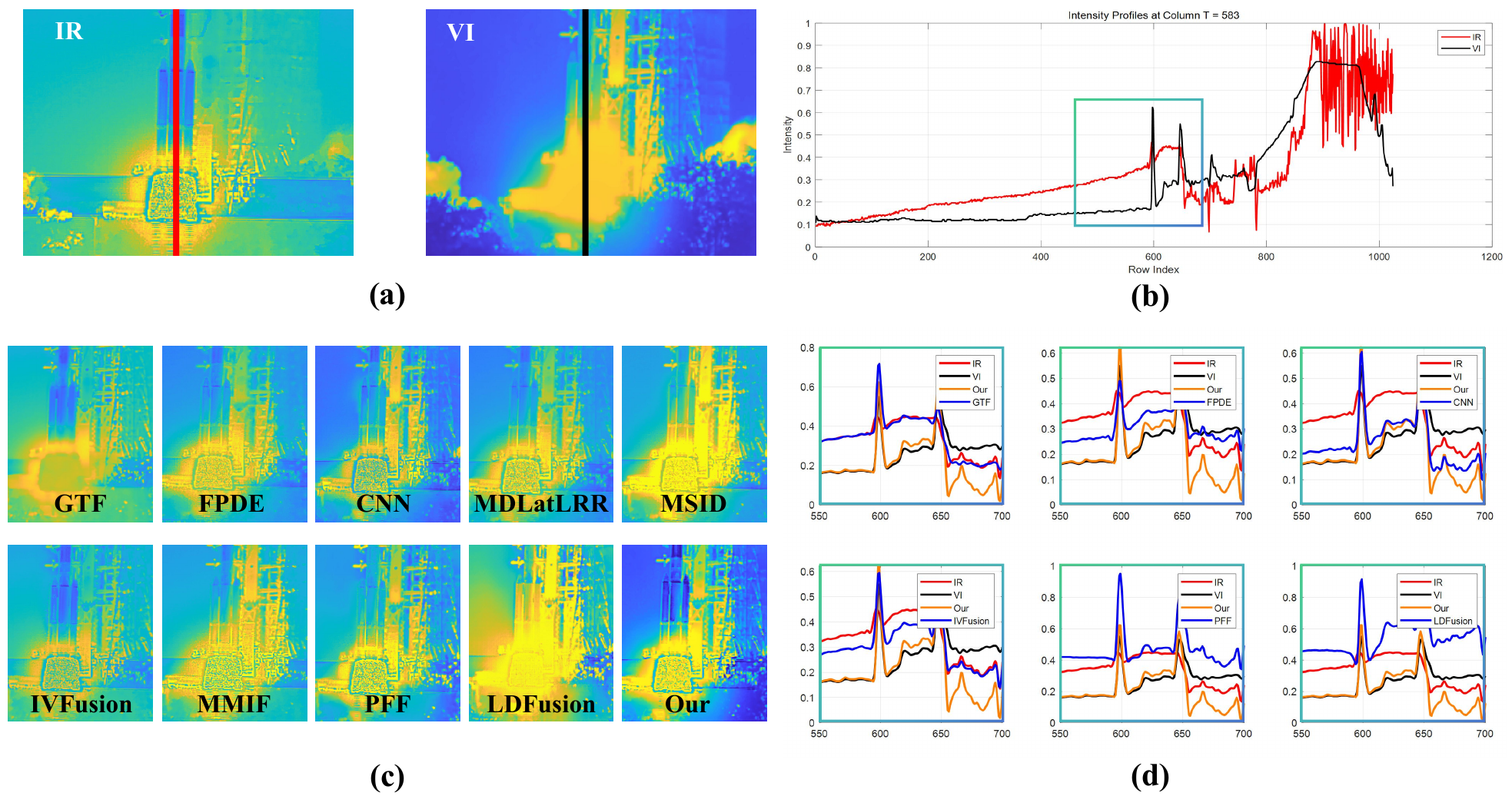}
		\caption{Qualitative analysis of the fusion result details. In the figure, (a) and (c) show grayscale mappings, where the yellow-to-blue gradient represents decreasing gray values. (a) Input image; (b) Gray value changes along the line in (a); (c) Region of interest in the results of various fusion methods; (d) Grayscale curves of different fusion results in the boxed area of (b).}
		\label{fig:precision}
	\end{figure*}
	
	\subsection{Fidelity Analysis}
	To assess the performance disparities among various image fusion techniques in terms of feature preservation, a comparative analysis was conducted utilizing gray-level curves. As depicted in Fig. \ref{fig:precision}, a notable divergence in gray-level values between the infrared (IR) and visible (VIS) images is evident. The gray-level mapping outcomes for the different fusion methods are shown in Fig. \ref{fig:precision}(c), where our method demonstrates superior performance in retaining information and preserving finer details.
	Further evaluation, shown in Fig. \ref{fig:precision}(d), compares several fusion methods with improved visual outcomes. Most existing methods (e.g., GTF, IVFusion, PFF, and LDFusion) fail to preserve visible information effectively, as their gray-level curves deviate significantly from the ideal reference curve. In contrast, both the CNN-based method and our approach exhibit superior performance, with our method displaying more pronounced fluctuations in the gray-level curve, which signifies enhanced image contrast.
	In conclusion, our fusion technique effectively integrates visible and infrared information, successfully preserving details and augmenting contrast. These improvements are crucial for enhancing feature extraction accuracy and facilitating subsequent analytical tasks.
	
\begin{figure}[t]
		\centering
		\includegraphics[width=0.4\textwidth]{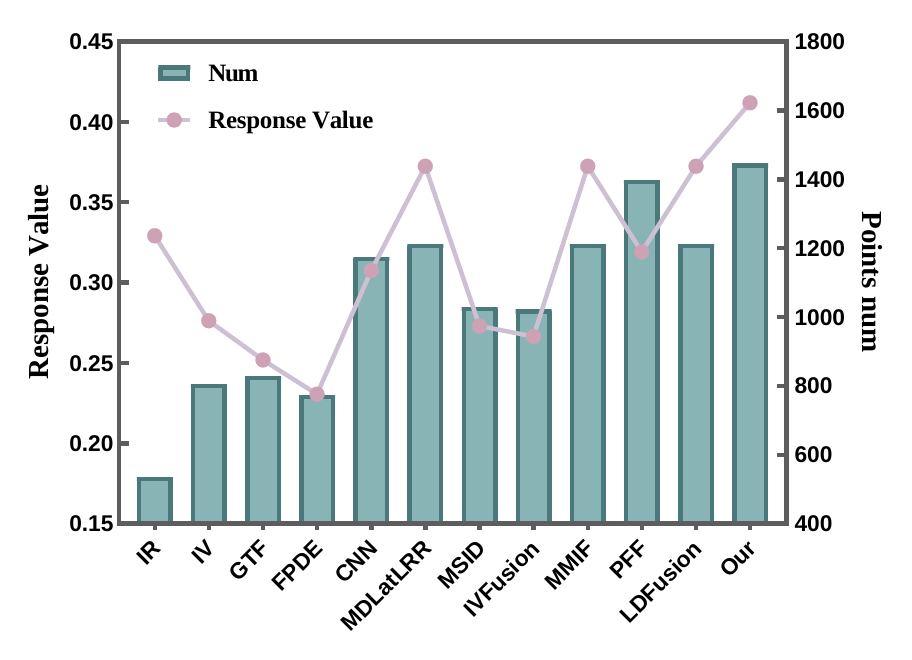}
		\caption{Quantitative results of feature points and response values in the region of interest for different fusion methods.}
		\label{fig:points_num}
	\end{figure}

	Fig. \ref{fig:points_num} shows feature extraction results in the region of interest using the standard method. Our approach yields more feature points with higher average response values, enhancing feature extraction quality and improving measurement accuracy. Although PFF generates more feature points, it exhibits lower response values due to the introduction of infrared noise and artifacts during fusion, as shown in Fig. \ref{fig:precision}(c).
	
	\subsection{Extended Application}
	The proposed method extends to various multi-modal fusion scenarios, including RGB-NIR, PET-MRI, and CT-MRI. The sequence in which images are fused significantly affects the results, so the input order is based on the information content of each image pair. The entropy value of each image is computed, with the higher-entropy image serving as the base and the lower-entropy one as the auxiliary input. This strategy optimizes information retention, yielding superior fusion results. The datasets used include VIS-NIR Scene\footnote{http://matthewalunbrown.com/nirscene/nirscene.html} (RGB-NIR) and Harvard\footnote{http://www.med.harvard.edu/AANLIB/home.html} (PET-MRI and CT-MRI). To evaluate our approach, IVFusion \cite{IVFusion} is used as a benchmark due to its extensive discussion in similar applications.
	
	\begin{figure*}[t]
		\centering
		\includegraphics[width=0.88\textwidth]{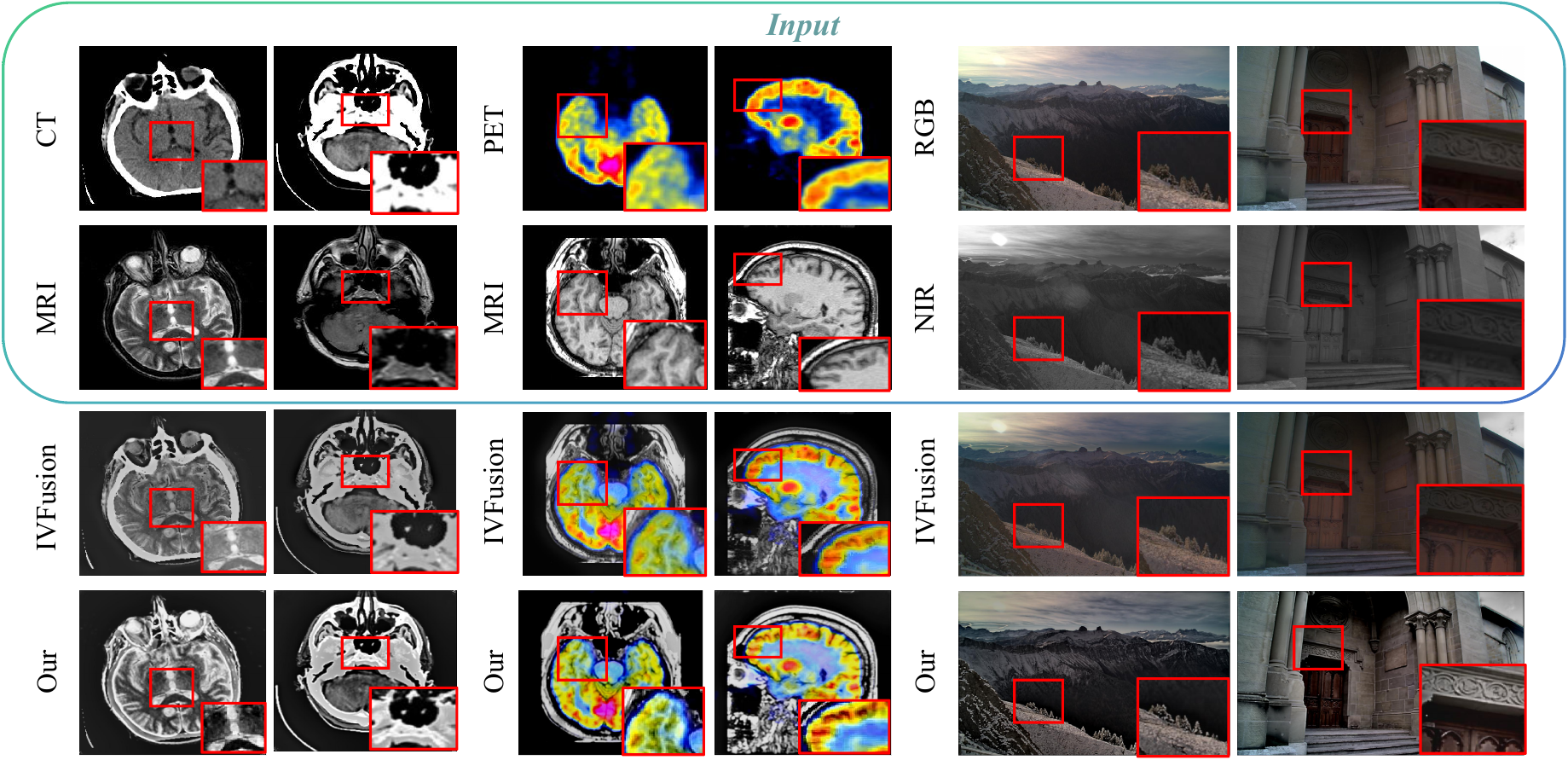}
		\caption{Our framework is applied to various scenarios, including CT-MRI image pairs, PET-MRI image pairs, and RGB-NIR image pairs. Each column represents a distinct group of image pairs.}
		\label{fig:app}
	\end{figure*}
	
    Fig. \ref{fig:app} presents the fusion outcomes across various datasets using different fusion methods. In the case of CT-MRI image pairs (highlighted in red), our method demonstrates more pronounced fusion, effectively revealing distinct textures and skeletal structures. For PET-MRI pairs, our approach excels in preserving texture more effectively than LVFusion, while maintaining critical brain information within the MRI images. Regarding RGB-NIR pairs, our fusion technique incorporates additional near-infrared (NIR) information into the RGB image, thereby enhancing details such as grass texture and door structure, as indicated by the red box. Overall, our method outperforms the other fusion methods across all multi-modal datasets.
	
	\section{Conclusion and Future Work}
	This paper presents an infrared-visible complementary fusion framework based on region perception, overcoming the limitations of traditional fusion methods that rely on single-exposure image pairs. By integrating novel sensing technologies, we establish an optimized synergy between multi-exposure and multi-modal fusion. Our framework consists of four key components: region perception model, feature adaptive merging, multi-exposure fusion, and infrared-visible complementary fusion. Through advancements in hardware and algorithms, the framework implements a three-stage computational workflow—multi-exposure fusion, multi-modal registration, and fusion—while ensuring measurement fidelity. Extensive evaluations on proprietary and benchmark datasets demonstrate the framework's superior fusion quality.
	Future work will focus on: 1) developing optimization strategies to improve computational efficiency without sacrificing performance, and 2) exploring integration with downstream measurement systems, especially for mission-critical applications like high-precision rocket trajectory monitoring and drift quantification.

	\section*{Acknowledgement}
	This work was supported by the Hunan Provincial Natural Science Foundation for Excellent Young Scholars (Grant No. 2023JJ20045), the Foundation of National Key Laboratory of Human Factors Engineering (Grant No. GJSD22006) , and the National Natural Science Foundation of China (Grant No. 12372189).

\section*{Data availability}

Data underlying the results presented in this paper are not publicly available at this time but may be obtained from the authors upon reasonable request.

\bibliographystyle{elsarticle-num-names} 
\bibliography{references.bib}

\end{document}